%% file: article.tex
\documentclass[11pt,a4paper]{article}

\usepackage[utf8]{inputenc}
\usepackage[T1]{fontenc}
\usepackage[english]{babel}
\usepackage{lmodern}
\usepackage[margin=1in]{geometry}
\usepackage{amsmath,amssymb}
\usepackage{graphicx}
\usepackage{booktabs}
\usepackage{enumitem}
\usepackage{listings}
\usepackage{xcolor}
\usepackage{microtype}
\usepackage{caption}
\usepackage{subcaption}
\usepackage{float}   
\usepackage[hidelinks,breaklinks]{hyperref}
\usepackage{url}

\definecolor{codekw}{rgb}{0.13,0.29,0.53}
\definecolor{codecom}{rgb}{0.45,0.45,0.45}

\lstdefinestyle{py}{
  language=Python,
  basicstyle=\ttfamily\footnotesize,
  keywordstyle=\color{codekw}\bfseries,
  stringstyle=\color{black},
  commentstyle=\color{codecom}\itshape,
  numbers=none,
  breaklines=true,
  breakatwhitespace=false,
  postbreak=\mbox{\textcolor{codecom}{$\hookrightarrow$}\space},
  showstringspaces=false,
  frame=none,
  framesep=4pt,
  columns=fullflexible,
  keepspaces=true,
  upquote=true,
  captionpos=b,
  aboveskip=8pt,
  belowskip=8pt,
}
\lstdefinestyle{plain}{
  basicstyle=\ttfamily\footnotesize,
  numbers=none,
  breaklines=true,
  breakatwhitespace=false,
  showstringspaces=false,
  frame=none,
  framesep=4pt,
  columns=fullflexible,
  keepspaces=true,
  upquote=true,
  captionpos=b,
}
\newcommand{\exlabel}[1]{\par\medskip\noindent Example~#1.\par\nobreak\smallskip}

\title{\bfseries QLoRA Fine-Tuning of Ministral LLM\\
       for Sequence-to-Function Protein Annotation}

\author{
  Demian Pavlyshenko\\
  \texttt{demkolviv@gmail.com}
  \and
  Bohdan Pavlyshenko\\
  \texttt{b.pavlyshenko@gmail.com}
}

\date{}

\begin{document}
\maketitle

\begin{abstract}
\noindent
Functional annotation of newly sequenced proteins remains a bottleneck in
molecular biology: the number of sequences in public repositories grows far
faster than the capacity for manual curation. Most computational approaches consider 
annotation as multi-label classification over a fixed ontology, 
 which constrains
predictions to a predefined label set.
In this work we study the
 the protein annotation as a sequence-to-text
generation problem. 
We fine-tune the 3B-parameter Ministral~3 base model with
QLoRA (4-bit NF4 quantization with low-rank adapters) on sequence annotation pairs. We assess predictions with an LLM-as-expert protocol: a GPT model prompted as a senior molecular-biology curator scores
organism identification as binary and function annotation quality. 
We conclude that QLoRA-fine-tuned compact LLMs can generate curator-style annotations with genuine biological value for a substantial subset of proteins. We also discuss future directions in data quality, model scaling, and evidence grounding that are needed to make the approach sufficiently reliable for practical use.
\end{abstract}


\vspace{1em}
{\hypersetup{linkcolor=black}
\setcounter{tocdepth}{2}   
\tableofcontents}
\vspace{1em}

\section{Introduction}
\label{sec:intro}

The gap between the number of known protein sequences and the number of
experimentally characterised proteins continues to widen. UniProtKB contains
hundreds of millions of entries, of which only a small, manually reviewed
fraction---the Swiss-Prot section---carries curated functional descriptions
written and evidence-tagged by expert biocurators \cite{uniprot2025}. Manual
curation is accurate but expensive and inherently unable to keep pace with
high-throughput sequencing, so the overwhelming majority of protein records rely
on automatic annotation transferred from homologs. Improving automated functional
annotation therefore remains one of the practically important problems of
computational biology.

The dominant computational formulation treats annotation as multi-label
classification over a controlled vocabulary, most often the Gene Ontology.
Homology-transfer baselines such as BLAST \cite{altschul1990blast} and
machine-learning systems such as DeepGOPlus \cite{kulmanov2020deepgoplus} assign
GO terms to sequences, and the CAFA challenges have established a standard
evaluation methodology for this setting \cite{zhou2019cafa3}. This formulation
has two limitations that motivate the present work. First, the prediction space
is closed: a classifier can only emit terms it was trained on, and cannot express
a functional nuance for which no term exists. Second, the output is a set of
identifiers rather than a description; the specific substrate, the interacting
partner, the direction of a regulatory effect or the physiological context---the
content that makes a Swiss-Prot \texttt{FUNCTION} comment useful to a
biologist---is either flattened into generic terms or lost entirely.

An alternative is to treat annotation as conditional text generation: given the
amino-acid sequence, generate the free-text functional description directly. This
formulation is open-ended, produces curator-style prose, and can be evaluated
against the reference annotation by biological meaning rather than by label
overlap. Recent multimodal systems such as Prot2Text \cite{abdine2024prot2text}
and Prot2Text-V2 \cite{prot2textv2} demonstrate that the formulation is viable,
typically by coupling a protein encoder (and, in the multimodal case, structural
information) to a text decoder. A natural question---and the one we address
here---is how much of this capability can be obtained from a \emph{general
purpose}, compact, openly available language model that receives nothing but the
raw amino-acid sequence as text, adapted with parameter-efficient fine-tuning on
commodity hardware.

This work makes the following contributions.

\begin{itemize}[leftmargin=1.4em,itemsep=3pt]
  \item We fine-tune the compact Ministral~3 3B base model
        \cite{ministral3b2512} with QLoRA \cite{dettmers2023qlora} to generate,
        from a spaced amino-acid sequence, the source organism and a UniProt-style
        \texttt{FUNCTION} description as a single JSON object
        (Section~\ref{sec:method}). The full run fits on one 80\,GB GPU and
        trains 135M adapter parameters, 3.39\% of the model.
  \item We propose and apply an expert-LLM evaluation protocol in which a GPT
        model prompted as a senior molecular biologist scores organism
        identification (binary) and functional quality (0--10) under an explicit
        rubric with structured-output decoding
        (Section~\ref{sec:method-eval}).
  \item We report quantitative results, training dynamics, worked examples and an
        error analysis on held-out proteins, and discuss the strengths and
        the limitations---in particular hallucination risk---of sequence-only
        generative annotation (Section~\ref{sec:results}).
\end{itemize}

All code excerpts required to reproduce the pipeline---dataset retrieval,
training configuration, generation and evaluation---are given in the appendices.

\section{Related Work}
\label{sec:related}

\subsection{Protein language models}
Self-supervised models trained on large protein-sequence corpora learn
representations that capture structural and functional signal without explicit
supervision. The ESM family established the paradigm at scale: ESM-1b showed that
biological structure and function emerge from masked-language-model pretraining on
250 million sequences \cite{rives2021esm1b}, and ESM-2 extended this to
atomic-level structure prediction directly from a language model
\cite{lin2023esm2}. ProtTrans systematically compared transformer architectures,
including ProtT5, trained on billions of protein tokens and showed that the
resulting embeddings are competitive with alignment-based features for per-residue
and per-protein tasks \cite{elnaggar2022prottrans}. Generative protein language
models such as ProGen \cite{madani2023progen} and ProtGPT2
\cite{ferruz2022protgpt2} model the sequence distribution itself and can generate
functional artificial proteins. These models treat amino-acid sequences as the
language to be modelled; the natural-language description of function is external
to them, and mapping embeddings to human-readable annotation requires an
additional supervised component.

\subsection{From labels to text: LLMs for protein annotation}
A growing body of work uses large language models to produce natural-language
descriptions of protein function rather than ontology labels. Prot2Text
\cite{abdine2024prot2text} was among the first to frame protein-function
prediction as free-text generation, combining a graph neural network over the
protein structure with a transformer decoder to generate Swiss-Prot-style
descriptions. Prot2Text-V2 \cite{prot2textv2} extends this line with a multimodal
contrastive alignment strategy: a protein encoder and a large instruction-tuned
language decoder are aligned so that sequence (and structure-derived)
representations are projected into the decoder's embedding space, improving the
biological faithfulness of generated descriptions; the 11B instruction-tuned
checkpoint is publicly distributed \cite{prot2textv2model}. Our setting is
deliberately more austere: we use no structural input, no protein encoder and no
contrastive alignment stage; the amino-acid sequence is presented to a general
purpose LLM as plain text, which isolates the contribution of instruction
fine-tuning itself and keeps the pipeline reproducible on a single GPU.

ProLLaMA \cite{lv2025prollama} takes a complementary route, converting a general
purpose LLaMA model into a multi-task protein language model through a two-stage
training procedure: continued pretraining on protein sequences to teach the model
"protein language", followed by instruction tuning that lets a single model handle
both controllable protein generation and protein-property prediction, with
low-rank adaptation used to keep training tractable; code and models are publicly
available \cite{prollamacode}. ProLLaMA supports our central premise---that a
general-purpose decoder-only LLM can be adapted to protein tasks with
parameter-efficient methods---while differing in objective: ProLLaMA focuses on
superfamily-conditioned sequence generation and property prediction, whereas we
target free-text curator-style functional annotation together with organism
identification.

\subsection{The Mistral and Ministral model families}
Mistral~7B \cite{jiang2023mistral7b} demonstrated that carefully trained compact
open-weight models can match or exceed considerably larger models, using grouped
query attention and sliding-window attention for efficient inference; Mixtral
\cite{jiang2024mixtral} extended the family with sparse mixture-of-experts
routing. The Ministral line targets the low-parameter, on-device regime: the
"Les Ministraux" models (3B and 8B) were introduced as efficient edge models with
strong knowledge and reasoning performance for their size
\cite{mistral2024ministraux}. In this work we use the Ministral~3 3B base
checkpoint \texttt{Ministral-3-3B-Base-2512} \cite{ministral3b2512}, a compact
decoder that loads in under 3\,GB of GPU memory in 4-bit precision, together with
the \texttt{mistral-common} tokenization and prompt-validation library
\cite{mistralcommon}. A base (non-instruction-tuned) checkpoint is a deliberate
choice: the target behaviour is a narrow, highly formatted annotation task, and
supervised fine-tuning defines the response format without competing with prior
instruction tuning.

\subsection{Parameter-efficient fine-tuning and 4-bit quantization}
Low-Rank Adaptation (LoRA) is a parameter-efficient fine-tuning method for large language models \cite{hu2022lora} that freezes pretrained weights and injects trainable low-rank
matrices into linear layers, so that for $W_0 \in \mathbb{R}^{d \times k}$ the
adapted transform is
\begin{equation}
  W_0 x + \Delta W x = W_0 x + \frac{\alpha}{r}\, B A x,
  \qquad B \in \mathbb{R}^{d \times r},\; A \in \mathbb{R}^{r \times k},\; r \ll \min(d,k),
  \label{eq:lora}
\end{equation}
reducing the number of trainable parameters by orders of magnitude while matching
full fine-tuning quality on many tasks. QLoRA \cite{dettmers2023qlora} makes this
practical for consumer and single-GPU settings by back-propagating through a
frozen base model quantized to 4-bit NormalFloat (NF4), a data type
information-theoretically suited to normally distributed weights, combined with
double quantization (quantizing the quantization constants themselves) and paged
optimizers that use NVIDIA unified memory to survive gradient-checkpointing memory
spikes. Together with block-wise 8-bit optimizer states
\cite{dettmers2022optimizers}, these techniques reduce the memory footprint of
fine-tuning a multi-billion-parameter model to a single commodity GPU. Our
implementation uses the Hugging Face Transformers \cite{wolf2020transformers} and
PEFT \cite{peft} libraries.

\subsection{Our previous work on fine-tuned Mistral models}
The present study continues a line of work on fine-tuning large language models
for domain-specific text analytics. In \cite{pavlyshenko2023disinformation}, a
Llama-family LLM was fine-tuned for the analysis of disinformation and fake news,
showing that a fine-tuned model can perform several analytic subtasks---fake-news
detection, extraction of named entities, and generation of structured analytic
summaries -- within a single generative interface, and that supervised fine-tuning
on a modest domain corpus substantially outperforms prompting alone. In
\cite{pavlyshenko2025crypto}, a fine-tuned Mistral model was combined with a
retrieval-augmented generation (RAG) pipeline for multilevel analysis of
cryptocurrency news, producing structured, multi-level analytic outputs (entity
and event extraction, sentiment and risk assessment, and higher-level summaries)
grounded in retrieved documents. Both studies use the same methodological
ingredients we apply here -- parameter-efficient fine-tuning of a compact
open-weight Mistral model, supervision by structured target texts, and evaluation
of generated free-text output. 

\subsection{LLM-based evaluation of generated text}
Evaluating generated annotations is itself a research problem. $n$-gram overlap
metrics such as BLEU \cite{papineni2002bleu} and ROUGE \cite{lin2004rouge} reward
surface similarity and are weakly correlated with biological correctness: a
prediction can paraphrase the reference perfectly and still name the wrong
substrate, or use entirely different wording for the same enzymatic activity.
Using strong LLMs as graders has been shown to correlate well with human judgment
for open-ended generation, both in the pairwise/absolute-score setting of MT-Bench
\cite{zheng2023judging} and in the rubric-driven G-Eval framework
\cite{liu2023geval}. We adopt this methodology with a domain-specific rubric
(Section~\ref{sec:method-eval}), which lets us separate organism identification
from functional correctness and to score partial biological credit explicitly.

\section{Dataset Construction}
\label{sec:dataset}

\subsection{Retrieval from UniProtKB/Swiss-Prot}
We constructed a protein sequence-to-function dataset from reviewed
UniProtKB/Swiss-Prot entries \cite{uniprot2025} retrieved through the UniProt
REST API \cite{ahmad2025uniprotapi}. Entries were selected with the query
\begin{center}
\texttt{reviewed:true AND length:[50 TO 1024]}
\end{center}
and requested in TSV format with the fields \texttt{accession},
\texttt{protein\_name}, \texttt{organism\_name}, \texttt{length},
\texttt{sequence}, \texttt{cc\_function}, \texttt{cc\_subcellular\_location},
\texttt{cc\_pathway}, \texttt{ec}, the three Gene Ontology aspects
(\texttt{go\_f}, \texttt{go\_p}, \texttt{go\_c}) and \texttt{xref\_interpro}.
Results were paginated through the API's \texttt{Link} header in batches of 500
entries. The \texttt{reviewed:true} filter restricts the corpus to manually
curated entries, so that every supervision target is a human-written functional
description rather than an automatically propagated one; the length window
excludes peptides too short to carry a reliable functional signal and sequences
too long to fit the training context economically. Protein sequences serve as
model inputs and the curated \texttt{cc\_function} comments serve as
natural-language supervision targets. The download code is given in
Appendix~\ref{app:dataset}, Listing~\ref{lst:download}.

\subsection{Filtering and construction of targets}
For the experiments reported here we retrieved 350{,}000 entries and applied the
following selection steps.

\begin{enumerate}[leftmargin=1.6em,itemsep=3pt]
  \item \textbf{Organism coverage.} We kept only entries whose organism occurs at
        least 500 times in the downloaded corpus. This yields \textbf{34
        organisms} and makes the organism-identification part of the task
        learnable: species that appear a handful of times cannot be inferred
        reliably from sequence alone, and their presence would add label noise
        without adding signal.
  \item \textbf{Length constraints.} We kept entries with sequence length
        $\leq 700$ residues and functional-comment length $\leq 850$ characters.
        The thresholds were chosen from the empirical quantiles of the two length
        distributions so as to retain the large majority of entries while bounding
        the token budget of both the prompt and the target.
  \item \textbf{Target assembly.} The supervision target concatenates the organism
        name and the curated functional comment,
        \begin{center}
        \texttt{target = Organism + "; " + Function [CC]}
        \end{center}
        so that a single generated string carries both predictions and can be
        split deterministically at the first semicolon at evaluation time. An
        example target is:
        \begin{quote}\small\sloppy
        \texttt{Oryctolagus cuniculus (Rabbit); FUNCTION: ATP-dependent RNA
        helicase which is a subunit of the eIF4F complex involved in cap
        recognition and is required for mRNA binding to ribosome. \dots}
        \end{quote}
\end{enumerate}

After these steps the dataset comprises 72{,}873 sequence--function
pairs. Immediately before training, three additional cleaning rules are applied
inside the training notebook: duplicate sequences are removed, sequences
containing non-standard amino-acid symbols (anything outside the 20-letter
alphabet \texttt{ACDEFGHIKLMNPQRSTVWY}) are discarded, and entries whose
functional comment matches \texttt{uncharacterized} or \texttt{function unknown}
are excluded, since such targets provide no biological supervision and teach the
model to produce evasive annotations. The resulting corpus  was
split randomly into a training set and a held-out test set  with test fraction 0.03. 

\section{Methodology}
\label{sec:method}

\subsection{Base model and quantization}
The base model is the Ministral~3 3B base checkpoint
\texttt{mistralai/Ministral-3-3B-Base-2512} \cite{ministral3b2512}, loaded with
4-bit NF4 quantization and double quantization following QLoRA
\cite{dettmers2023qlora}, with bfloat16 as the compute dtype:

\begin{center}\small
\texttt{load\_in\_4bit=True}, \texttt{bnb\_4bit\_quant\_type="nf4"},
\texttt{bnb\_4bit\_use\_double\_quant=True},
\texttt{bnb\_4bit\_compute\_dtype=torch.bfloat16}.
\end{center}

In this configuration the frozen base weights occupy approximately 2.6\,GB of GPU
memory. Tokenization uses the model's native Tekken tokenizer through the
\texttt{mistral-common} backend \cite{mistralcommon}, which also renders the
instruction template.

\subsection{Sequence representation and prompt format}
\label{sec:method-prompt}
Amino-acid sequences are presented to the model as plain text, with a single
space inserted between consecutive residues (\texttt{"M K A V S P \dots"}). This
whitespace separation prevents the subword tokenizer from merging residues into
arbitrary multi-character tokens that vary with local context: spacing makes the
mapping from residues to tokens essentially one-to-one and position-stable, at the
cost of a longer prompt. The sequence is wrapped in \texttt{<protein>} tags inside
an instruction that fixes the task and the output format, and the whole is
rendered through the Mistral chat template, giving prompts of the form
\texttt{<s>[INST]\dots[/INST]}:

\begin{lstlisting}[style=plain,caption={Instruction prompt used for fine-tuning and inference (the model continues after \texttt{[/INST]}).},label={lst:prompt}]
<s>[INST]You are a protein-function annotation assistant.

Infer only what is supported by the amino-acid sequence.
Return concise valid JSON with one key: "function".

Protein sequence:
<protein>
S R D N G P D G M E P E G V I E S N W N E I V D S F D D M N L S E S L L R G I Y A Y G F E K ...
</protein>
[/INST]
\end{lstlisting}

The training target is the reference annotation serialised as a one-key JSON
object, terminated by the end-of-sequence token:

\begin{lstlisting}[style=plain,caption={Training target: organism and curated \texttt{FUNCTION} comment inside a JSON envelope.},label={lst:target}]
{"function": "Oryctolagus cuniculus (Rabbit); FUNCTION: ATP-dependent RNA helicase which is a
subunit of the eIF4F complex involved in cap recognition and is required for mRNA binding to
ribosome. ..."}</s>
\end{lstlisting}

The JSON envelope makes the output machine-parseable and gives the model an
unambiguous stopping condition; the organism prefix before the first semicolon
makes taxonomic prediction an explicit, separately scorable part of the task.

Sequences are built by concatenating the tokenized prompt with the tokenized
target rather than by rendering a two-turn conversation. This is both a
correctness and a practical decision: it guarantees that the prompt tokens are an
exact prefix of the training sequence -- which the loss mask relies on -- and it
avoids the serving-mode validation in \texttt{mistral-common}, which requires the
last message of a conversation to come from the user. Loss is computed only over
the answer tokens: label positions corresponding to the prompt are set to $-100$
(the ignore index), so the model is never trained to reproduce the instruction or
the amino-acid sequence. Examples are truncated to 1{,}536 tokens.
Listing~\ref{lst:tokenize} in Appendix~\ref{app:training} gives the
implementation.

\subsection{QLoRA configuration}
\label{sec:method-qlora}
Low-rank adapters (Equation~\ref{eq:lora}) are injected into all seven projection
matrices of every transformer block -- the attention projections
\texttt{q\_proj}, \texttt{k\_proj}, \texttt{v\_proj}, \texttt{o\_proj} and the
feed-forward projections \texttt{gate\_proj}, \texttt{up\_proj},
\texttt{down\_proj} -- with rank $r = 64$, scaling $\alpha = 128$ (so
$\alpha/r = 2$) and adapter dropout 0.05; no bias terms are trained. This yields
135{,}004{,}160 trainable parameters out of 3{,}984{,}094{,}208, i.e.\
\textbf{3.39\%} of the model. Before adapter injection the quantized model is
prepared for $k$-bit training (input embeddings cast appropriately, layer norms
kept in higher precision, gradient checkpointing enabled with non-reentrant
checkpointing).

\subsection{Training setup}
\label{sec:method-training}
Training ran for 3 epochs on a single NVIDIA A100-SXM4-80GB GPU with bfloat16
compute, a per-device batch of 16 and 2 gradient-accumulation steps (effective
batch 32), for a total of 6{,}336 optimizer steps. We used a learning rate of
$1\times10^{-4}$ with a cosine schedule and 1\% warmup, the paged 8-bit AdamW
optimizer \cite{dettmers2022optimizers,dettmers2023qlora}, gradient checkpointing,
and length-grouped batching so that sequences of similar length are batched
together and padding overhead is reduced. Batches are padded to a multiple of
eight tokens with the EOS token, padding positions are masked out of both the
attention mask and the loss. Validation loss was computed every 300 steps on a
fixed 300-example subset of the held-out split, and checkpoints were written to
persistent storage at the same interval so that an interrupted session could be
resumed. The complete configuration is listed in Table~\ref{tab:hyper} and in
Appendix~\ref{app:training}, Listing~\ref{lst:training}.

\begin{table}[H]
\centering
\small
\caption{Model, QLoRA and optimization hyperparameters.}
\label{tab:hyper}
\begin{tabular}{ll}
\toprule
\textbf{Setting} & \textbf{Value} \\
\midrule
Base model & \texttt{mistralai/Ministral-3-3B-Base-2512} \\
Total / trainable parameters & 3{,}984{,}094{,}208 / 135{,}004{,}160 (3.39\%) \\
Quantization & 4-bit NF4, double quantization, bfloat16 compute \\
LoRA rank $r$ / $\alpha$ / dropout & 64 / 128 / 0.05 \\
LoRA target modules & q,k,v,o\_proj; gate,up,down\_proj \\
Max sequence length & 1{,}536 tokens \\
Epochs / optimizer steps & 3 / 6{,}336 \\
Batch size (device $\times$ accumulation) & $16 \times 2 = 32$ \\
Learning rate / schedule / warmup & $1\times10^{-4}$ / cosine / 1\% \\
Optimizer & paged AdamW, 8-bit states \\
Precision & bf16 (TF32 matmul enabled) \\
Gradient checkpointing & enabled (non-reentrant) \\
Batching strategy & group-by-length, pad to multiple of 8 \\
Hardware / wall-clock time & 1 $\times$ A100-SXM4-80GB / 7.18 h \\
Throughput & 7.84 samples/s \\
\bottomrule
\end{tabular}
\end{table}

\subsection{Inference}
At inference time the adapter is attached to the same 4-bit base model, ensuring
that the frozen weights the adapter sees are bit-identical to those it was trained
against. Annotations are produced by greedy decoding (\texttt{do\_sample=False}),
which is appropriate for an extraction-like task where a single most-probable
annotation is wanted rather than diversity. Prompts within a batch are left-padded
so that every sequence ends exactly at the generation boundary, the key-value
cache is re-enabled (it is disabled during gradient-checkpointed training), and
generation stops at the EOS token that terminates the JSON object. We used a
generation batch size of 16.

\subsection{Expert-LLM evaluation protocol}
\label{sec:method-eval}
Because surface-overlap metrics do not measure biological correctness, we evaluate
generated annotations with an LLM-as-expert protocol
\cite{zheng2023judging,liu2023geval} implemented against the GPT API with the
model \texttt{gpt-5.6-terra}. The evaluation pipeline proceeds as follows.

\begin{enumerate}[leftmargin=1.6em,itemsep=3pt]
  \item \textbf{Parsing.} Reference and prediction strings are split at the first
        semicolon into an organism part and a function part, the \texttt{FUNCTION:}
        prefix is stripped, and UniProt evidence tags (\texttt{\{ECO:\dots\}},
        \texttt{(PubMed:\dots)}, \texttt{(By similarity)}) are removed. Evidence
        tags are curation metadata; leaving them in would let the grader reward
        the imitation of citation formatting.
  \item \textbf{Grading prompt.} The system prompt casts the model as a senior
        expert in molecular biology, biochemistry, enzymology and protein science
        familiar with Swiss-Prot curation practice, and defines three tasks:
        (A) \texttt{organism\_predicted} $\in \{0,1\}$, where 1 requires the same
        species -- formatting, common names, synonyms and strain differences are
        ignored, while a different species, even a close relative, scores 0;
        (B) \texttt{function\_quality} $\in \{0,\dots,10\}$ under an explicit
        rubric that weighs molecular function first, biological process second and
        specific details third, with anchors from 10 (biologically equivalent) to
        0 (wrong, contradictory or empty); and (C) \texttt{comments}, a short
        expert remark on what the prediction got right or wrong. The rubric
        instructs the grader to ignore wording and citation style, not to reward
        length, to penalise unsupported added claims, to cap vague
        "may be involved in" predictions at 3, and to grade function independently
        of organism correctness. The full prompt is reproduced in
        Appendix~\ref{app:eval}.
  \item \textbf{Structured decoding.} The grader returns a schema-constrained
        object with a free-text rationale preceding each score
        \cite{openai2026api}, so that scores are always integers in the admissible
        range and the rationale is available for audit. Requiring the rationale
        before the score follows the chain-of-thought-style evaluation used in
        G-Eval \cite{liu2023geval}.
  \item \textbf{Execution.} Requests run in parallel with retry and back-off, and
        every graded row is appended to a cache file, so an interrupted evaluation
        resumes without re-grading. Results are written as a CSV with the columns
        \texttt{reference}, \texttt{prediction}, \texttt{organism\_predicted},
        \texttt{function\_quality} and \texttt{comments}.
\end{enumerate}

We report organism accuracy (the mean of \texttt{organism\_predicted}), the mean
and median function-quality score, and the share of predictions scoring $\geq 7$,
which we treat as the threshold for an annotation that a curator could use as a
starting point.

\section{Results and Discussion}
\label{sec:results}

\subsection{Training dynamics}
\begin{figure}[!htb]
\centering
\begin{subfigure}[b]{0.48\textwidth}
  \includegraphics[width=\textwidth]{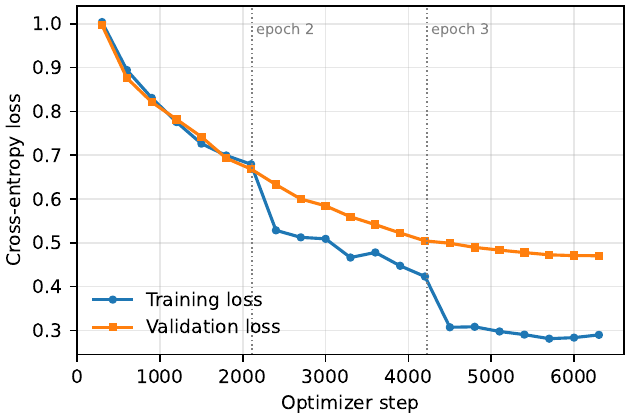}
  \caption{Training and validation loss.}
  \label{fig:loss}
\end{subfigure}
\hfill
\begin{subfigure}[b]{0.48\textwidth}
  \includegraphics[width=\textwidth]{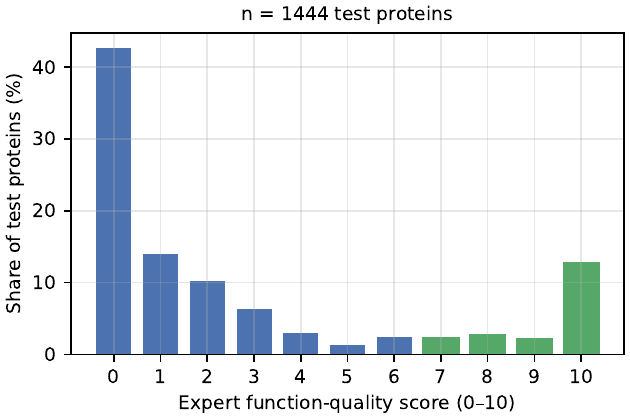}
  \caption{Expert function-quality scores; green bars mark usable
           annotations (score $\geq 7$).}
  \label{fig:scores}
\end{subfigure}
\caption{Training dynamics (left) and the distribution of expert scores on the
         held-out test proteins (right).}
\label{fig:main}
\end{figure}

Figure~\ref{fig:loss} shows training and validation cross-entropy over the 6{,}336
optimizer steps. Both curves decrease monotonically, from 1.004/0.997
(training/validation) at step 300 to 0.290/0.470 at the end of training; the
overall training loss averaged over the run is 0.536. The characteristic
step-shaped drops at the epoch boundaries (steps 2{,}112 and 4{,}224) reflect the
model re-encountering training examples, and the widening gap between the two
curves in the third epoch -- validation loss flattens at $\approx 0.470$ while
training loss continues to fall to $\approx 0.290$ -- indicates the onset of
memorisation. Validation loss does not increase, so the run was not stopped early,
but the flattening suggests that further epochs on this dataset would mostly
increase the train--validation gap; additional data, rather than additional
epochs, is the direction with headroom. Figure~\ref{fig:scores} shows 
expert function-quality scores.

\subsection{Quantitative evaluation}
We generated annotations for 1{,}444 held-out test proteins and graded each one
with the protocol of Section~\ref{sec:method-eval}. The headline results are
summarised in Table~\ref{tab:results}: the fine-tuned model identifies the source
organism correctly for \textbf{40.4\%} of test proteins, and its functional
annotations receive a mean expert score of \textbf{2.72/10}.

\begin{table}[H]
\centering
\small
\caption{Expert-LLM evaluation of the fine-tuned Ministral~3 3B model on 1{,}444
         held-out test proteins. Graded by \texttt{gpt-5.6-terra} under the rubric
         of Appendix~\ref{app:eval}.}
\label{tab:results}
\begin{tabular}{lr}
\toprule
\textbf{Metric} & \textbf{Value} \\
\midrule
Organism accuracy (\texttt{organism\_predicted} $=1$)      & 40.4\% \\
Function quality, mean                                      & 2.72 / 10 \\
Function quality, median                                    & 1.0 \\
Function quality $\geq 7$ (usable annotation)               & 19.9\% \\
Function quality $\geq 8$                                   & 17.7\% \\
Function quality $= 10$ (biologically equivalent)           & 12.7\% \\
Function quality $\leq 1$                                   & 56.6\% \\
Organism correct \emph{and} function quality $\geq 7$       & 8.7\% \\
\bottomrule
\end{tabular}
\end{table}

 Two modes dominate: 42.6\% of predictions receive a
score of 0 and 12.7\% receive a full 10, with comparatively few predictions in the
middle of the scale (scores 4--6 together account for only 6.9\%). In other words,
the model does not usually produce mediocre approximations of the correct
annotation; it either recovers the curated function almost exactly or produces a
fluent annotation for a different protein. This is consistent with a
retrieval-like behaviour induced by fine-tuning: when the test sequence is close
to something the model has seen, it reproduces the corresponding curated
description with high fidelity; otherwise it falls back on a plausible but
unsupported annotation.

Organism prediction and function quality are only weakly coupled. Among
predictions that score 10 on function, the organism is correct in just 50.0\% of
cases, against 34.5\% among those scoring 0; mean function quality is 2.96 when
the organism is correct and 2.56 when it is not. The joint success rate -- correct
organism \emph{and} function score $\geq 7$ -- is 8.7\%. The frequent combination
of a perfect functional description with the wrong species is instructive: highly
conserved proteins have near-identical curated annotations across organisms, so
the model can reproduce the functional text from a homolog while assigning it to
the species that dominates the training data for that protein family. Example~1
in Appendix~\ref{app:examples} is exactly this case: the pig cornified
envelope protein is annotated with the correct function verbatim but assigned to
human.

\subsection{Qualitative analysis of high-scoring predictions}
Appendix~\ref{app:examples} reproduces the graded examples for the two upper
score bands. The score-10 cases (Appendix~\ref{app:ex10}) share a clear profile:
the predicted text reproduces the molecular activity, the biological process and
the specific details of the curated comment -- for example a RAB5-specific GTPase
activating protein that explicitly excludes activity on RAB4 and RAB11, or an
SSU-processome component described with its correct ribosome-biogenesis
context -- so that only the wording, and frequently the organism, differs from the
reference. These are cases in which sequence-only annotation genuinely works: the
model has internalised a family-to-function mapping precise enough to reconstruct
curator-level detail.

A recurring defect is nevertheless visible even in this best band: the
\emph{evidence tags} that the model reproduces alongside the annotation text are
almost never the reference ones. UniProt terminates a \texttt{FUNCTION} comment
with machine-readable provenance such as
\texttt{\{ECO:\allowbreak 0000269\textbar\allowbreak PubMed:\allowbreak 12214233\}}, where the ECO code states the
type of evidence and the PubMed identifier points to the publication supporting
the statement. Having been trained on targets that end with such tags, the model
emits them as part of the expected output format: 87.0\% of predictions carry an
ECO tag, matching the 88.2\% of references that do, and 31.3\% cite at least one
PubMed identifier. Those identifiers, however, are essentially never the correct
ones -- of the predictions that cite literature, 96.9\% cite at least one PubMed
identifier absent from the reference annotation, and only 3.5\% share any
identifier with it. The defect is independent of functional accuracy: among
score-10 predictions that carry citations, 80.0\% still cite at least one
identifier that the reference does not. In the worked examples this is visible
throughout Appendix~\ref{app:ex10}: seven of the ten score-10 predictions attach
evidence tags that differ from the reference, for instance a reference supported
by \texttt{\{ECO:\allowbreak 0000269\textbar\allowbreak PubMed:\allowbreak 12214233\}} against a prediction claiming
inference by similarity, or a prediction citing three PubMed identifiers none of
which appears in the curated entry. The explanation is structural: PubMed
identifiers are arbitrary integers with no relation to the amino-acid sequence, so
they cannot be inferred from the input at all; the model reproduces the
\emph{shape} of a citation, which is precisely the definition of a fabricated
reference. Our grading rubric instructs the expert model to ignore evidence tags,
so this defect does not affect the reported scores -- but it must be stripped
before any downstream use, and it is a reminder that fluent formatting is
generated independently of factual grounding.

The score-7 cases (Appendix~\ref{app:ex7}) are equally informative because they
expose the characteristic partial-credit failure modes: (i) \emph{correct family,
missing specifics} -- a metallothionein is correctly identified as a
cysteine-rich heavy-metal binder, but the prediction omits the Cu/Zn chelation and
the role in copper remobilisation during leaf senescence and seed development;
(ii) \emph{correct molecular identity, generic biology} -- a Frizzled-family ligand
is recognised, but the kidney-specific pronephric developmental role and its
Notch-pathway context are replaced by generic developmental language; and (iii)
\emph{correct core function with an unsupported addition}, where the model adds a
specific claim (for instance that a Complex I subunit is essential for complex
assembly) that the curated entry does not support. The third mode is the most
dangerous in practice, since fluent, specific and false statements are exactly
what an annotation consumer is least able to detect.

\subsection{Error analysis}
Inspecting the graded rationales for low-scoring predictions reveals four recurring
error types.

\begin{itemize}[leftmargin=1.4em,itemsep=3pt]
  \item \textbf{Confident misassignment (dominant).} The prediction is a
        well-formed, internally consistent annotation for a \emph{different}
        protein family. As one grader comment puts it: "Both the species and
        protein assignment are incorrect. The predicted synaptotagmin-like calcium
        sensor is unrelated to the reference CDKN1B Thr-187 phosphatase." The
        output looks exactly like a good annotation and can only be recognised as
        wrong by a domain expert or an independent method.
  \item \textbf{Unsupported specificity.} The core function is right but the model
        adds substrate, partner or mechanism details that the curated entry does
        not contain -- a direct consequence of training on rich, specific targets:
        the model learns that annotations \emph{sound} specific and supplies
        specificity whether or not the sequence justifies it.
  \item \textbf{Generic under-specification.} Vague predictions ("May be involved
        in the regulation of cell proliferation") that the rubric caps at a low
        score. These are the model's fallback when no strong family signal is
        detected, and they are the least harmful failure mode.
  \item \textbf{Organism bias.} Errors skew towards the organisms best represented
        in the training corpus, human above all; the model uses the taxonomic
        prior of its training distribution rather than species-specific sequence
        features. Predicted functional texts are also shorter than the curated
        references (median 143 vs.\ 220 characters), reflecting a bias towards the
        shorter, more generic portion of the annotation distribution.
\end{itemize}

\subsection{Discussion}
\label{sec:discussion}

\paragraph{What the results support.}
The central positive finding is that a 3B-parameter general-purpose model, given
nothing but a spaced amino-acid string and adapted with 3.39\% trainable
parameters on a single GPU for about seven hours, produces annotations that a
strict expert rubric judges biologically equivalent to Swiss-Prot curation for
one prediction in eight, and usable for one in five. No protein encoder, no
structural input, no multiple-sequence alignment and no ontology were involved.
This is a strong statement about the efficiency of QLoRA adaptation: the
combination of 4-bit NF4 quantization, low-rank adapters and paged 8-bit
optimizers \cite{dettmers2023qlora,dettmers2022optimizers} moves a task that would
otherwise require multi-GPU full fine-tuning into a setting reproducible by a
single researcher, which matters for the reproducibility of domain-specific LLM
research generally, as in our earlier applications to news analytics
\cite{pavlyshenko2023disinformation,pavlyshenko2025crypto}.

\paragraph{Limitations of sequence-only annotation.}
The results equally delimit the approach. Protein function is determined by
three-dimensional structure, binding interfaces and cellular context; a decoder-only
LLM reading residues as text has access to none of these directly and must infer
family membership from sequence patterns alone. Systems that inject structural or
encoder-derived representations -- Prot2Text \cite{abdine2024prot2text},
Prot2Text-V2 \cite{prot2textv2} -- address precisely this gap, and the comparison
suggests the natural next step for our pipeline is to condition on embeddings from
a dedicated protein language model \cite{lin2023esm2,elnaggar2022prottrans} or on
domain signatures such as InterPro \cite{blum2021interpro} rather than on raw
residues. A second, methodological limitation is that our random split does not
control for homology between training and test proteins: some part of the
high-scoring tail reflects near-duplicate family coverage rather than
generalisation to genuinely novel proteins, and a sequence-identity-clustered
split would give a stricter estimate of generalisation.

\paragraph{Hallucination risk.}
The failure profile described above is the principal obstacle to practical
deployment. Unlike a classifier, which expresses uncertainty through low
probabilities over a known label set, a generative annotator expresses uncertainty
by producing a confident annotation for a different protein. The model provides no
calibrated signal distinguishing its score-10 from its score-0 outputs, and the
fluency of the output is uncorrelated with its correctness. Any practical use
therefore requires an external verification layer: agreement with homology search,
consistency with domain signatures, self-consistency across sampled generations,
or a curator in the loop. Generated annotations should be treated as curation
\emph{drafts}, never as evidence, and should never be deposited into a database
without experimental or at least computational corroboration; propagating
hallucinated annotations into public resources would degrade exactly the curated
corpora such models are trained on.

The fabricated evidence tags discussed above sharpen this point. The model
reproduces UniProt provenance syntax -- ECO evidence codes and
\texttt{PubMed:\textit{nnnnnnnn}} identifiers -- with the statistics of the training
distribution (87.0\% of predictions carry an ECO tag) while the identifiers
themselves are almost always wrong (96.9\% of citing predictions cite at least one
identifier absent from the reference), \emph{including} in predictions whose
biology is judged perfect. This is the clearest available demonstration that
surface form and factual grounding are generated by the same mechanism and
therefore carry no mutual guarantee: a fabricated citation is easy to detect
automatically, by checking the identifier against PubMed, whereas a fabricated
substrate or interaction partner is not. Practically, evidence tags should be
removed from the training targets altogether -- so that the model is never taught
to emit provenance it cannot possess -- and re-attached, if at all, by an external
retrieval step that links the generated statement to real literature.

\paragraph{Applicability to uncharacterized proteins.}
The intended use case -- proteins without curated annotation -- is also where our
evidence is weakest: we deliberately removed uncharacterized entries from
training, and our test set consists of proteins that \emph{do} have curated
annotations. Since performance is driven by proximity to well-annotated families,
performance on genuinely novel sequences is expected to be lower than the numbers
reported here, and establishing it requires prospective evaluation against
subsequently curated entries or experimental validation.

\paragraph{On ablations.}
We report a single end-to-end configuration ($r=64$, $\alpha=128$). A systematic
ablation of the adapter rank (e.g.\ $r=32$ vs.\ $r=64$), of $\alpha/r$, of the set
of adapted projections and of the sequence representation (spaced vs.\ unspaced
residues) would isolate their individual contributions, but each variant requires
a full training run plus a graded evaluation pass, which we leave to future work
rather than report untested. We note only that the rank chosen here is at the
upper end of typical QLoRA settings, consistent with the fact that the target
task -- mapping residue patterns to biological text -- is further from the base
model's pretraining distribution than a typical natural-language fine-tune.

\paragraph{On the evaluation protocol.}
Grading with an LLM introduces its own uncertainty: the grader may be biased
toward fluent text, may apply the rubric inconsistently across biological
subdomains, and cannot verify claims against primary literature. We mitigate this
with an explicit rubric, mandatory rationales, structured decoding and a
deterministic split of the two subtasks, and we release the per-example rationales
for audit. Validating the grader against expert human annotators on a sample
remains necessary future work; the protocol should be read as a scalable proxy for
expert review \cite{zheng2023judging,liu2023geval}, not as a replacement for it.

\section{Conclusion}
\label{sec:conclusion}

We tested a QLoRA approach for fine-tuning a compact Ministral large language
model for protein functional annotation, formulating the problem as
sequence-to-text generation: the model reads a raw amino-acid sequence and writes
the source organism together with a free-text, curator-style \texttt{FUNCTION}
description. For evaluation we used expert estimation performed by another large
language model -- a GPT API model prompted as a senior molecular biologist -- which
scores organism identification and functional correctness separately under an
explicit rubric.

The results show that for a substantial share of protein sequences the approach
yields very good annotations and correct organism detection: 12.7\% of held-out
predictions were judged biologically equivalent to the curated Swiss-Prot
annotation, 19.9\% reached a usable quality level (score $\geq 7$), and the source
organism was identified correctly for 40.4\% of proteins -- using a 3B-parameter
model adapted on a single GPU with 3.39\% of its parameters trainable. The score
distribution is bimodal: the model tends either to recover the curated function
almost exactly or to produce a fluent annotation belonging to a different protein,
which makes hallucination, rather than vagueness, the characteristic failure mode
and mandates external verification before any practical use. The same mechanism
produces fabricated provenance: the model reproduces the UniProt evidence-tag
syntax of its training targets -- ECO codes and PubMed identifiers such as
\texttt{\{ECO:\allowbreak 0000269\textbar\allowbreak PubMed:\allowbreak 12214233\}} -- but the cited identifiers are
almost never those of the reference entry, and this holds even for annotations
judged biologically equivalent to the curated text. Evidence tags should
therefore be stripped from the supervision targets in future iterations, since
literature provenance cannot be inferred from an amino-acid sequence and can only
be supplied by an external retrieval step.

It is expected that with larger training datasets, optimized QLoRA
hyperparameters, or larger Mistral-family models, the obtained results can be
essentially improved. Beyond scale, the most promising directions are to ground
generation in additional evidence -- structural information, embeddings from
dedicated protein language models, or InterPro/Pfam domain signatures supplied in
the prompt -- to add calibrated confidence estimation so that unreliable
annotations can be filtered automatically, to adopt homology-aware train/test
splits for a stricter measure of generalisation, and to validate the LLM grader
against human expert annotators. 

\bibliographystyle{unsrt}
\bibliography{references}

\appendix
\newpage

\section{Appendix}
\label{app}

\subsection{Dataset Retrieval and Preprocessing}
\label{app:dataset}

Entries were retrieved from the UniProt REST API \cite{ahmad2025uniprotapi} on
2026-09-19 using the current UniProtKB/Swiss-Prot release \cite{uniprot2025}.
Listing~\ref{lst:download} shows the paginated download, and
Listing~\ref{lst:filters} the filtering and target construction.

\begin{lstlisting}[caption={Paginated retrieval of reviewed Swiss-Prot entries from the UniProt REST API.},label={lst:download}]
import re, requests, pandas as pd
from io import StringIO

url = "https://rest.uniprot.org/uniprotkb/search"

params = {
    # Reviewed (manually curated) entries only; length window in residues.
    "query": "reviewed:true AND length:[50 TO 1024]",
    "format": "tsv",
    "fields": ",".join([
        "accession", "protein_name", "organism_name", "length", "sequence",
        "cc_function",              # curated FUNCTION comment = supervision target
        "cc_subcellular_location", "cc_pathway", "ec",
        "go_f", "go_p", "go_c",     # Gene Ontology aspects (not used for training)
        "xref_interpro",            # domain signatures (reserved for future work)
    ]),
    "size": 500,                    # entries per page
}

def get_next_link(response):
    """UniProt paginates through the HTTP Link header: <url>; rel="next"."""
    match = re.search(r'<([^>]+)>;\s*rel="next"', response.headers.get("Link", ""))
    return match.group(1) if match else None

pages, total_records = [], 0
next_url, next_params = url, params.copy()

while next_url and total_records < 350_000:
    response = requests.get(next_url, params=next_params, timeout=60)
    response.raise_for_status()

    page = pd.read_csv(StringIO(response.text), sep="\t")
    pages.append(page)
    total_records += len(page)

    next_url = get_next_link(response)
    next_params = None              # later URLs already contain all parameters
    print(f"Downloaded: {total_records:,} proteins")

df = pd.concat(pages, ignore_index=True)
\end{lstlisting}

\begin{lstlisting}[caption={Organism and length filters, and construction of the supervision target.},label={lst:filters}]
# Keep organisms that are represented well enough to be learnable (>= 500 entries).
organism_list = df["Organism"].value_counts()
organism_list = organism_list[organism_list >= 500].index.tolist()
len(organism_list)                       # -> 34 organisms

# Bound the token budget of prompt and target.
df["seq_len"]  = df["Sequence"].str.len()
df["func_len"] = df["Function [CC]"].str.len()
df = df[(df.seq_len <= 700) & (df.func_len <= 850)]

df1 = df[df.Organism.isin(organism_list)]        # -> 72,873 entries

# Target = organism, then the curated FUNCTION comment, joined by "; ".
# Evaluation splits the generated string back at the first semicolon.
df1["target"] = df1["Organism"] + "; " + df1["Function [CC]"]

df1[["Sequence", "target"]].to_csv("prot_seq5.csv", index=False)
\end{lstlisting}

\begin{lstlisting}[caption={Cleaning rules and train/test split applied before training.},label={lst:clean}]
train_df = (
    df.rename(columns={"Sequence": "sequence", "target": "function"})
      [["sequence", "function"]].dropna()
      .drop_duplicates(subset=["sequence"])          # no duplicated sequences
)

# Standard 20-letter amino-acid alphabet only (no X, B, Z, U, O).
train_df = train_df[train_df["sequence"].str.fullmatch(r"[ACDEFGHIKLMNPQRSTVWY]+")]

# Drop targets that carry no biological supervision.
train_df = train_df[~train_df["function"].str.contains(
    r"uncharacterized|function unknown", case=False, na=False)]

train_df = train_df[train_df["sequence"].str.len().between(50, 700)]

dataset = Dataset.from_pandas(train_df, preserve_index=False)
dataset = dataset.train_test_split(test_size=0.03, seed=42)
# -> train: 67,553 examples, test: 2,090 examples
\end{lstlisting}

\newpage
\subsection{Training Configuration}
\label{app:training}

\begin{lstlisting}[caption={Prompt construction, target serialisation and loss masking. Only the answer tokens contribute to the loss.},label={lst:tokenize}]
def space_amino_acids(sequence: str) -> str:
    # One space between residues: keeps the residue-to-token mapping stable.
    return " ".join(sequence)

def build_user_text(sequence: str) -> str:
    return f"""You are a protein-function annotation assistant.

Infer only what is supported by the amino-acid sequence.
Return concise valid JSON with one key: "function".

Protein sequence:
<protein>
{space_amino_acids(sequence)}
</protein>
"""

def build_target(function: str) -> str:
    # "<organism>; FUNCTION: ..." wrapped in a one-key JSON envelope.
    return json.dumps({"function": function.strip()}, ensure_ascii=False)

def tokenize_example(example):
    user_message = {"role": "user",
                    "content": [{"type": "text",
                                 "text": build_user_text(example["sequence"])}]}

    # Prompt ending exactly where the assistant answer starts: <s>[INST] ... [/INST]
    # (mistral-common validates in "serving" mode, where the last message must be
    #  the user's, so the assistant turn is never passed to the chat template.)
    prompt_ids = tokenizer.apply_chat_template(
        [user_message], add_generation_prompt=True, tokenize=True, return_dict=True,
    )["input_ids"]

    # Expected answer + EOS, exactly as the template renders an assistant turn.
    answer_ids = tokenizer.encode(
        build_target(example["function"]), add_special_tokens=False,
    ) + [tokenizer.eos_token_id]

    input_ids = (prompt_ids + answer_ids)[:MAX_TOKENS]   # MAX_TOKENS = 1536
    prompt_length = min(len(prompt_ids), len(input_ids))

    # -100 = ignore index: train only on the assistant's annotation.
    labels = [-100] * prompt_length + input_ids[prompt_length:]

    return {"input_ids": input_ids,
            "attention_mask": [1] * len(input_ids),
            "labels": labels}
\end{lstlisting}

\begin{lstlisting}[caption={4-bit base model, QLoRA adapters and training arguments.},label={lst:training}]
# --- 4-bit NF4 base model (QLoRA) ------------------------------------------
bnb_config = BitsAndBytesConfig(
    load_in_4bit=True,
    bnb_4bit_quant_type="nf4",              # NormalFloat4, information-optimal for
    bnb_4bit_use_double_quant=True,         # normally distributed weights
    bnb_4bit_compute_dtype=torch.bfloat16,
)

model = Mistral3ForConditionalGeneration.from_pretrained(
    "mistralai/Ministral-3-3B-Base-2512",
    quantization_config=bnb_config, dtype=torch.bfloat16, device_map={"": 0},
)
model.config.use_cache = False               # incompatible with gradient checkpointing

# --- LoRA adapters on all projection matrices ------------------------------
model = prepare_model_for_kbit_training(
    model, use_gradient_checkpointing=True,
    gradient_checkpointing_kwargs={"use_reentrant": False},
)

lora_config = LoraConfig(
    r=64, lora_alpha=128, lora_dropout=0.05, bias="none", task_type="CAUSAL_LM",
    target_modules=["q_proj", "k_proj", "v_proj", "o_proj",
                    "gate_proj", "up_proj", "down_proj"],
)
model = get_peft_model(model, lora_config)
model.print_trainable_parameters()
# trainable params: 135,004,160 || all params: 3,984,094,208 || trainable%: 3.3886

# --- Optimization ----------------------------------------------------------
training_args = TrainingArguments(
    output_dir=OUTPUT_DIR,
    num_train_epochs=3,
    learning_rate=1e-4, lr_scheduler_type="cosine", warmup_steps=0.01,
    per_device_train_batch_size=16, gradient_accumulation_steps=2,   # effective 32
    per_device_eval_batch_size=16,
    train_sampling_strategy="group_by_length",   # less padding within a batch
    gradient_checkpointing=True,
    gradient_checkpointing_kwargs={"use_reentrant": False},
    bf16=True, tf32=True,
    optim="paged_adamw_8bit",                    # paged optimizer, 8-bit states
    logging_steps=50, eval_strategy="steps", eval_steps=300,
    save_strategy="steps", save_steps=300, save_total_limit=2,
    remove_unused_columns=False, report_to="none", seed=42,
)

trainer = Trainer(model=model, args=training_args,
                  train_dataset=tokenized["train"], eval_dataset=eval_subset,
                  data_collator=ProteinDataCollator())
trainer.train()
\end{lstlisting}

\begin{lstlisting}[caption={Batched greedy generation with left padding, used to annotate the test proteins.},label={lst:generate}]
@torch.inference_mode()
def generate_batch(sequences: list[str]) -> list[str]:
    prompts = [prompt_ids_for(s) for s in sequences]
    max_len = max(len(x) for x in prompts)
    pad_id = tokenizer.eos_token_id

    # Left padding: every prompt ends exactly where generation starts.
    input_ids = torch.tensor([[pad_id] * (max_len - len(x)) + x for x in prompts],
                             device=model.device)
    attention_mask = torch.tensor([[0] * (max_len - len(x)) + [1] * len(x)
                                   for x in prompts], device=model.device)

    output = model.generate(
        input_ids=input_ids, attention_mask=attention_mask,
        max_new_tokens=MAX_NEW_TOKENS, do_sample=False,   # greedy decoding
        use_cache=True, pad_token_id=pad_id, eos_token_id=pad_id,
    )
    return [tokenizer.decode(row[max_len:].tolist(), skip_special_tokens=True).strip()
            for row in output]
\end{lstlisting}

\newpage
\subsection{Expert Evaluation Prompt and Pipeline}
\label{app:eval}

The system prompt below is used verbatim with the \texttt{gpt-5.6-terra} model;
the user message contains the parsed reference and predicted organism and function
for one protein.

\begin{lstlisting}[style=plain,caption={System prompt defining the expert grader, the two scoring tasks and the comment task.},label={lst:evalprompt}]
You are a senior expert in molecular biology, biochemistry, enzymology and protein
science, with deep familiarity with UniProtKB/Swiss-Prot curation practice and
biological taxonomy.

You evaluate a model that reads a protein amino-acid sequence and predicts
(1) the source organism and (2) the protein's function, in UniProt "FUNCTION" style.
You compare each prediction to the curated reference annotation for the same protein.
You judge biological meaning, not wording. Be rigorous, consistent and strict.

TASK A: organism_predicted (0 or 1)
- 1: the predicted organism is the same species as the reference. Ignore formatting,
  common names in parentheses, synonyms, and strain, isolate or subspecies differences
  within the same species (e.g. "Escherichia coli (strain K12)" vs
  "Escherichia coli O157:H7" -> 1).
- 0: a different species (even a close relative, e.g. Mus musculus vs Rattus
  norvegicus -> 0), a genus only, or a missing organism.

TASK B: function_quality (integer 0-10)
Compare the predicted function to the reference function. Weigh, in order of importance:
  1. Molecular function / biochemical activity (catalytic activity, reaction, substrate,
     binding partner or ligand, protein family or class).
  2. Biological process and pathway (the role the protein plays in the cell or organism).
  3. Specific details: subcellular context, regulation, complex membership, key partners,
     cofactors, specificity.
Scoring anchors:
  10  Biologically equivalent: same activity, process and key specifics; only wording differs.
  8-9 Same molecular function and process; minor details missing or slightly imprecise.
  6-7 Correct protein family / general activity and broadly correct role, but important
      specifics (substrate, partner, pathway) missing or partially wrong.
  4-5 Partially correct: the right general functional area (e.g. "kinase",
      "transcription factor", "transporter") but the specific role differs or is wrong.
  2-3 Mostly incorrect; only a weak or generic overlap (e.g. both "involved in development").
  1   Essentially unrelated function, but still a plausible protein function.
  0   Completely wrong, contradictory, empty, or nonsense.
Rules:
- Ignore evidence tags, PubMed IDs, citations and the style of the text.
- Do not reward length. Extra claims that are wrong or not supported by the reference
  lower the score. Correct but more specific detail is acceptable.
- A vague prediction ("May be involved in...") scores at most 3 unless the reference
  is equally vague and matches it.
- If the prediction is truncated, judge only the content present.
- Grade the function independently of whether the organism is correct.

TASK C: comments
Write a short expert comment (1-2 sentences, at most ~40 words) on the LLM's
annotation as a whole: what it got right, what it got wrong or missed, and any
notable issue (hallucinated specifics, wrong family, too vague, truncated,
correct organism but wrong protein, etc.).

Give a brief rationale for each task before the score.
\end{lstlisting}

\begin{lstlisting}[caption={Parsing of annotations, structured-output schema and the graded API call.},label={lst:evalcode}]
# --- Split "Organism; FUNCTION: ..." and strip curation metadata ------------
EVIDENCE_RE = re.compile(r"\{ECO:[^}]*\}|\((?:PubMed|By similarity)[^)]*\)")

def split_annotation(text: str) -> tuple[str, str]:
    organism, sep, function = text.partition(";")
    if not sep:
        return "", text.strip()
    function = re.sub(r"^\s*FUNCTION:\s*", "", function.strip())
    function = re.sub(r"\s{2,}", " ", EVIDENCE_RE.sub("", function)).strip()
    return organism.strip(), function

# --- Schema-constrained grader output --------------------------------------
class Evaluation(BaseModel):
    organism_rationale: str          # rationale precedes the score (G-Eval style)
    organism_predicted: Literal[0, 1]
    function_rationale: str
    function_quality: int = Field(ge=0, le=10)
    comments: str

def evaluate_row(row) -> dict:
    user_prompt = USER_TEMPLATE.format(
        ref_organism=row.ref_organism, ref_function=row.ref_function,
        pred_organism=row.pred_organism, pred_function=row.pred_function,
    )
    response = client.responses.parse(
        model="gpt-5.6-terra",
        input=[{"role": "system", "content": SYSTEM_PROMPT},
               {"role": "user",   "content": user_prompt}],
        text_format=Evaluation,      # structured output: scores always in range
    )
    return response.output_parsed.model_dump()

# Graded rows are cached to JSONL, so an interrupted evaluation resumes;
# requests run in a thread pool with exponential back-off on failure.
\end{lstlisting}

\newpage
\subsection{Graded Prediction Examples}
\label{app:examples}

The examples below are verbatim outputs of the fine-tuned model on held-out test
proteins, with the reference annotation and the expert grader's comment. Evidence
tags are shown as generated, which makes a defect of the model directly visible:
even among the score-10 predictions of Appendix~\ref{app:ex10}---whose functional
text is judged biologically equivalent to the curated annotation---seven of ten
carry provenance tags that differ from the reference. Example~7 replaces the
curated support \texttt{\{ECO:\allowbreak 0000269\textbar\allowbreak PubMed:\allowbreak 12214233\}} with a claim of
inference by similarity, and Example~6 cites three PubMed identifiers, none of
which appears in the curated entry. PubMed identifiers are arbitrary integers
unrelated to the amino-acid sequence and thus cannot be predicted from it; the
model reproduces the \emph{form} of a citation rather than its content. The
grading rubric of Appendix~\ref{app:eval} instructs the expert model to ignore
evidence tags, so this does not influence the scores reported in
Section~\ref{sec:results}.

\subsubsection{Predictions scoring 10}
\label{app:ex10}
{\sloppy
\input{examples_score10}
\par}

\newpage
\subsubsection{Predictions scoring 7}
\label{app:ex7}
{\sloppy
\input{examples_score7}
\par}

\end{document}

%% file: examples_score10.tex
\exlabel{1}
\begin{description}[leftmargin=1.9em,style=nextline,itemsep=1pt,topsep=2pt,font=\normalfont]
\item[Reference:] Sus scrofa (Pig); FUNCTION: Cross-linked envelope protein of keratinocytes. It is a keratinocyte protein that first appears in the cell cytosol, but ultimately becomes cross-linked to membrane proteins by transglutaminase. All that results in the formation of an insoluble envelope beneath the plasma membrane.
\item[Prediction:] Homo sapiens (Human); FUNCTION: Cross-linked envelope protein of keratinocytes. It is a keratinocyte protein that first appears in the cell cytosol, but ultimately becomes cross-linked to membrane proteins by transglutaminase. All that results in the formation of an insoluble envelope beneath the plasma membrane.
\item[Expert comment:] Function annotation matches the curated role essentially exactly. The organism assignment is incorrect: human was predicted instead of pig.
\end{description}

\exlabel{2}
\begin{description}[leftmargin=1.9em,style=nextline,itemsep=1pt,topsep=2pt,font=\normalfont]
\item[Reference:] Homo sapiens (Human); FUNCTION: Acts as a GTPase activating protein for RAB5. Does not act on RAB4 or RAB11 (By similarity). \{ECO:\allowbreak{}0000250|\allowbreak{}UniProtKB:\allowbreak{}Q8IZP1\}.
\item[Prediction:] Homo sapiens (Human); FUNCTION: Acts as a GTPase activating protein for RAB5. Does not act on RAB4 or RAB11 (By similarity). \{ECO:\allowbreak{}0000250\}.
\item[Expert comment:] Exact match to the curated organism and functional annotation, including Rab substrate specificity.
\end{description}

\exlabel{3}
\begin{description}[leftmargin=1.9em,style=nextline,itemsep=1pt,topsep=2pt,font=\normalfont]
\item[Reference:] Neurospora crassa (strain ATCC 24698 / 74-OR23-1A / CBS 708.71 / DSM 1257 / FGSC 987); FUNCTION: Component of the ribosome, a large ribonucleoprotein complex responsible for the synthesis of proteins in the cell. The small ribosomal subunit (SSU) binds messenger RNAs (mRNAs) and translates the encoded message by selecting cognate aminoacyl-transfer RNA (tRNA) molecules. The large subunit (LSU) contains the ribosomal catalytic site termed the peptidyl transferase center (PTC), which catalyzes the formation of peptide bonds, thereby polymerizing the amino acids delivered by tRNAs into a polypeptide chain. The nascent polypeptides leave the ribosome through a tunnel in the LSU and interact with protein factors that function in enzymatic processing, targeting, and the membrane insertion of nascent chains at the exit of the ribosomal tunnel. \{ECO:\allowbreak{}0000305|\allowbreak{}PubMed:\allowbreak{}34815343\}.
\item[Prediction:] Saccharomyces cerevisiae (strain ATCC 204508 / S288c) (Baker's yeast); FUNCTION: Component of the ribosome, a large ribonucleoprotein complex responsible for the synthesis of proteins in the cell. The small ribosomal subunit (SSU) binds messenger RNAs (mRNAs) and translates the encoded message by selecting cognate aminoacyl-transfer RNA (tRNA) molecules. The large subunit (LSU) contains the ribosomal catalytic site termed the peptidyl transferase center (PTC), which catalyzes the formation of peptide bonds, thereby polymerizing the amino acids delivered by tRNAs into a polypeptide chain. The nascent polypeptides leave the ribosome through a tunnel in the LSU and interact with protein factors that function in enzymatic processing, targeting, and the membrane insertion of nascent chains at the exit of the ribosomal tunnel. \{ECO:\allowbreak{}0000305|\allowbreak{}PubMed:\allowbreak{}22096102\}.
\item[Expert comment:] The ribosome-component annotation matches the reference essentially exactly. The organism assignment is incorrect: baker's yeast is not Neurospora crassa.
\end{description}

\exlabel{4}
\begin{description}[leftmargin=1.9em,style=nextline,itemsep=1pt,topsep=2pt,font=\normalfont]
\item[Reference:] Oryctolagus cuniculus (Rabbit); FUNCTION: Receptor for adenosine. The activity of this receptor is mediated by G proteins which activate adenylyl cyclase (By similarity). \{ECO:\allowbreak{}0000250\}.
\item[Prediction:] Mus musculus (Mouse); FUNCTION: Receptor for adenosine. The activity of this receptor is mediated by G proteins which activate adenylyl cyclase.
\item[Expert comment:] The molecular function and signaling mechanism are correct. However, the predicted organism is mouse rather than rabbit.
\end{description}

\exlabel{5}
\begin{description}[leftmargin=1.9em,style=nextline,itemsep=1pt,topsep=2pt,font=\normalfont]
\item[Reference:] Drosophila melanogaster (Fruit fly); FUNCTION: Odorant receptor which mediates acceptance or avoidance behavior, depending on its substrates. The odorant receptor repertoire encodes a large collection of odor stimuli that vary widely in identity, intensity, and duration. May form a complex with Orco to form odorant-sensing units, providing sensitive and prolonged odorant signaling and calcium permeability. \{ECO:\allowbreak{}0000269|\allowbreak{}PubMed:\allowbreak{}16615896\}.
\item[Prediction:] Drosophila melanogaster (Fruit fly); FUNCTION: Odorant receptor which mediates acceptance or avoidance behavior, depending on its substrates. The odorant receptor repertoire encodes a large collection of odor stimuli that vary widely in identity, intensity, and duration. May form a complex with Orco to form odorant-sensing units, providing sensitive and prolonged odorant signaling and calcium permeability. Involved in the behavioral responses to ethyl acetate, pentyl acetate, and anisole. \{ECO:\allowbreak{}0000269|\allowbreak{}PubMed:\allowbreak{}21613503, ECO:\allowbreak{}0000269|\allowbreak{}PubMed:\allowbreak{}21857978\}.
\item[Expert comment:] The annotation is functionally equivalent to the curated odorant-receptor description and correctly identifies the species. It additionally provides plausible odorant-response specificity without introducing a contradiction.
\end{description}

\exlabel{6}
\begin{description}[leftmargin=1.9em,style=nextline,itemsep=1pt,topsep=2pt,font=\normalfont]
\item[Reference:] Saccharomyces cerevisiae (strain ATCC 204508 / S288c) (Baker's yeast); FUNCTION: Component of the kinetochore, a multiprotein complex that assembles on centromeric DNA and attaches chromosomes to spindle microtubules, mediating chromosome segregation and sister chromatid segregation during meiosis and mitosis. Component of the inner kinetochore constitutive centromere-associated network (CCAN), which serves as a structural platform for outer kinetochore assembly. \{ECO:\allowbreak{}0000269|\allowbreak{}PubMed:\allowbreak{}9987135\}.
\item[Prediction:] Saccharomyces cerevisiae (strain ATCC 204508 / S288c) (Baker's yeast); FUNCTION: Component of the kinetochore, a multiprotein complex that assembles on centromeric DNA and attaches chromosomes to spindle microtubules, mediating chromosome segregation and sister chromatid segregation during meiosis and mitosis. Component of the inner kinetochore constitutive centromere-associated network (CCAN), which serves as a structural platform for outer kinetochore assembly. \{ECO:\allowbreak{}0000269|\allowbreak{}PubMed:\allowbreak{}11266451, ECO:\allowbreak{}0000269|\allowbreak{}PubMed:\allowbreak{}12514103, ECO:\allowbreak{}0000269|\allowbreak{}PubMed:\allowbreak{}12514104\}.
\item[Expert comment:] Exact match to the curated organism and functional annotation, including CCAN/inner kinetochore localization and the chromosome-segregation role.
\end{description}

\exlabel{7}
\begin{description}[leftmargin=1.9em,style=nextline,itemsep=1pt,topsep=2pt,font=\normalfont]
\item[Reference:] Mus musculus (Mouse); FUNCTION: Putative pheromone receptor implicated in the regulation of social and reproductive behavior. \{ECO:\allowbreak{}0000269|\allowbreak{}PubMed:\allowbreak{}12214233\}.
\item[Prediction:] Homo sapiens (Human); FUNCTION: Putative pheromone receptor implicated in the regulation of social and reproductive behavior. \{ECO:\allowbreak{}0000250|\allowbreak{}UniProtKB:\allowbreak{}Q8VIC6\}.
\item[Expert comment:] Function annotation is biologically equivalent to the reference, but the organism is incorrect: human rather than mouse.
\end{description}

\exlabel{8}
\begin{description}[leftmargin=1.9em,style=nextline,itemsep=1pt,topsep=2pt,font=\normalfont]
\item[Reference:] Homo sapiens (Human); FUNCTION: Odorant receptor. \{ECO:\allowbreak{}0000305\}.
\item[Prediction:] Homo sapiens (Human); FUNCTION: Odorant receptor. \{ECO:\allowbreak{}0000305\}.
\item[Expert comment:] Both organism and function match the curated annotation exactly.
\end{description}

\exlabel{9}
\begin{description}[leftmargin=1.9em,style=nextline,itemsep=1pt,topsep=2pt,font=\normalfont]
\item[Reference:] Gallus gallus (Chicken); FUNCTION: Pseudokinase which, in complex with CAB39/MO25 (CAB39/MO25alpha or CAB39L/MO25beta), binds to and activates STK11/LKB1. Adopts a closed conformation typical of active protein kinases and binds STK11/LKB1 as a pseudosubstrate, promoting conformational change of STK11/LKB1 in an active conformation (By similarity). \{ECO:\allowbreak{}0000250\}.
\item[Prediction:] Xenopus laevis (African clawed frog); FUNCTION: Pseudokinase which, in complex with CAB39/MO25 (CAB39/MO25alpha or CAB39L/MO25beta), binds to and activates STK11/LKB1. Adopts a closed conformation typical of active protein kinases and binds STK11/LKB1 as a pseudosubstrate, promoting conformational change of STK11/LKB1 in an active conformation (By similarity). \{ECO:\allowbreak{}0000250\}.
\item[Expert comment:] Function annotation is fully concordant with the reference, including the CAB39/MO25-dependent LKB1 activation mechanism. The organism assignment is incorrect: African clawed frog rather than chicken.
\end{description}

\exlabel{10}
\begin{description}[leftmargin=1.9em,style=nextline,itemsep=1pt,topsep=2pt,font=\normalfont]
\item[Reference:] Saccharomyces cerevisiae (strain ATCC 204508 / S288c) (Baker's yeast); FUNCTION: Component of the mitochondrial ribosome (mitoribosome), a dedicated translation machinery responsible for the synthesis of mitochondrial genome-encoded proteins, including at least some of the essential transmembrane subunits of the mitochondrial respiratory chain. The mitoribosomes are attached to the mitochondrial inner membrane and translation products are cotranslationally integrated into the membrane. \{ECO:\allowbreak{}0000305|\allowbreak{}PubMed:\allowbreak{}25609543, ECO:\allowbreak{}0000305|\allowbreak{}PubMed:\allowbreak{}28154081\}.
\item[Prediction:] Saccharomyces cerevisiae (strain ATCC 204508 / S288c) (Baker's yeast); FUNCTION: Component of the mitochondrial ribosome (mitoribosome), a dedicated translation machinery responsible for the synthesis of mitochondrial genome-encoded proteins, including at least some of the essential transmembrane subunits of the mitochondrial respiratory chain. The mitoribosomes are attached to the mitochondrial inner membrane and translation products are cotranslationally integrated into the membrane. \{ECO:\allowbreak{}0000305|\allowbreak{}PubMed:\allowbreak{}24675956, ECO:\allowbreak{}0000305|\allowbreak{}PubMed:\allowbreak{}25609543\}.
\item[Expert comment:] Exact match for both organism and function. The annotation accurately captures mitoribosome membership and its membrane-associated mitochondrial translation role.
\end{description}

%% file: examples_score7.tex
\exlabel{11}
\begin{description}[leftmargin=1.9em,style=nextline,itemsep=1pt,topsep=2pt,font=\normalfont]
\item[Reference:] Homo sapiens (Human); FUNCTION: May regulate calcium-dependent activities in the endoplasmic reticulum lumen or post-ER compartment.
\item[Prediction:] Bos taurus (Bovine); FUNCTION: Involved in regulation of vitamin K-dependent carboxylation of multiple N-terminal glutamate residues. Seems to inhibit gamma-carboxylase GGCX. Binds 7 calcium ions with a low affinity (By similarity). \{ECO:\allowbreak{}0000250\}.
\item[Expert comment:] The organism is incorrect. The functional prediction broadly fits an ER calcium-dependent regulatory role, but adds specific gamma-carboxylase inhibition and calcium stoichiometry not supported by the curated reference.
\end{description}

\exlabel{12}
\begin{description}[leftmargin=1.9em,style=nextline,itemsep=1pt,topsep=2pt,font=\normalfont]
\item[Reference:] Arabidopsis thaliana (Mouse-ear cress); FUNCTION: Repressor of jasmonate responses. Jasmonoyl-isoleucine (JA-Ile) specifically promotes COI1-TIFY10A/JAZ1 interaction. Interacts with COI1 and inositol pentakisphosphate to form a high-affinity jasmonates coreceptor. \{ECO:\allowbreak{}0000269|\allowbreak{}PubMed:\allowbreak{}17637675, ECO:\allowbreak{}0000269|\allowbreak{}PubMed:\allowbreak{}17637677, ECO:\allowbreak{}0000269|\allowbreak{}PubMed:\allowbreak{}19151223\}.
\item[Prediction:] Arabidopsis thaliana (Mouse-ear cress); FUNCTION: Repressor of jasmonate responses. \{ECO:\allowbreak{}0000269|\allowbreak{}PubMed:\allowbreak{}19151223\}.
\item[Expert comment:] Correct species and central repressor role. The annotation is substantially incomplete because it lacks the JA-Ile-specific COI1 interaction and coreceptor mechanism.
\end{description}

\exlabel{13}
\begin{description}[leftmargin=1.9em,style=nextline,itemsep=1pt,topsep=2pt,font=\normalfont]
\item[Reference:] Arabidopsis thaliana (Mouse-ear cress); FUNCTION: Involved in multiple distinct defense responses. May function as a disease resistance (R) protein. \{ECO:\allowbreak{}0000269|\allowbreak{}PubMed:\allowbreak{}14756307, ECO:\allowbreak{}0000269|\allowbreak{}PubMed:\allowbreak{}15604743\}.
\item[Prediction:] Arabidopsis thaliana (Mouse-ear cress); FUNCTION: May be involved in pectin and/or xylans biosynthesis in cell walls. \{ECO:\allowbreak{}0000250\}.
\item[Expert comment:] Correct species, DNA-binding specificity, and general activator role. The predicted target-gene context is incorrect and misses the defining MEIS1-dependent activation of PF4.
\end{description}

\exlabel{14}
\begin{description}[leftmargin=1.9em,style=nextline,itemsep=1pt,topsep=2pt,font=\normalfont]
\item[Reference:] Homo sapiens (Human); FUNCTION: Inhibits signal transduction by increasing the GTPase activity of G protein alpha subunits thereby driving them into their inactive GDP-bound form. Binds to G(i)-alpha and G(o)-alpha, but not to G(s)-alpha (By similarity). \{ECO:\allowbreak{}0000250\}.
\item[Prediction:] Homo sapiens (Human); FUNCTION: Inhibits signal transduction by increasing the GTPase activity of G protein alpha subunits thereby driving them into their inactive GDP-bound form. Binds to G(i) alpha-1, G(i) alpha-2, G(i) alpha-3 and G(q) alpha.
\item[Expert comment:] Correct human assignment and correct RGS/GTPase-accelerating activity. Binding specificity is materially inaccurate: G(o)-alpha is omitted and unsupported G(q)-alpha binding is asserted.
\end{description}

\exlabel{15}
\begin{description}[leftmargin=1.9em,style=nextline,itemsep=1pt,topsep=2pt,font=\normalfont]
\item[Reference:] Arabidopsis thaliana (Mouse-ear cress); FUNCTION: Required for constitutive and alternative pre-mRNA splicing.
\item[Prediction:] Arabidopsis thaliana (Mouse-ear cress); FUNCTION: Probably involved in intron recognition and spliceosome assembly.
\item[Expert comment:] Correct species and correctly identifies a spliceosomal/pre-mRNA splicing role. It misses the specific requirement in both constitutive and alternative splicing and adds unverified mechanistic detail.
\end{description}

\exlabel{16}
\begin{description}[leftmargin=1.9em,style=nextline,itemsep=1pt,topsep=2pt,font=\normalfont]
\item[Reference:] Arabidopsis thaliana (Mouse-ear cress); FUNCTION: Transcription factor. Involved in responses to transient and long-term shade. Required for the light-mediated inhibition of hypocotyl elongation. Necessary for rapid light-induced expression of the photomorphogenesis- and circadian-related gene APRR9. Seems to play a role in multiple PHYB responses, such as flowering transition and petiole elongation. \{ECO:\allowbreak{}0000269|\allowbreak{}PubMed:\allowbreak{}14668869, ECO:\allowbreak{}0000269|\allowbreak{}PubMed:\allowbreak{}16565297, ECO:\allowbreak{}0000269|\allowbreak{}PubMed:\allowbreak{}16891401\}.
\item[Prediction:] Arabidopsis thaliana (Mouse-ear cress); FUNCTION: Transcription factor that binds specifically to a 5'-AA[AG]G-3' consensus core sequence. \{ECO:\allowbreak{}0000250\}.
\item[Expert comment:] Correct species and core transcription-factor function. The annotation misses the defining PHYB/shade-response and photomorphogenic roles, while adding an unverified sequence-specific DNA-binding claim.
\end{description}

\exlabel{17}
\begin{description}[leftmargin=1.9em,style=nextline,itemsep=1pt,topsep=2pt,font=\normalfont]
\item[Reference:] Xenopus laevis (African clawed frog); FUNCTION: Growth factor that plays an essential role in the regulation of embryonic development, cell proliferation, cell migration, survival and chemotaxis. Potent mitogen for cells of mesenchymal origin. Signaling is modulated by the formation of heterodimers with PDGFB (By similarity). \{ECO:\allowbreak{}0000250\}.
\item[Prediction:] Danio rerio (Zebrafish) (Brachydanio rerio); FUNCTION: Growth factor that plays an essential role in the regulation of embryonic development, cell proliferation, cell migration, survival and chemotaxis. Potent mitogen for cells of mesenchymal origin. Required for normal proliferation and recruitment of pericytes and vascular smooth muscle cells in the central nervous system, skin, lung, heart and placenta. Required for normal blood vessel development, and for normal development of kidney glomeruli. Plays an important role in wound healing. Signaling is modulated by the formation of heterodimers with PDGFA (By similarity). \{ECO:\allowbreak{}0000250\}.
\item[Expert comment:] The core PDGF-like growth-factor activity and developmental role are correct, but the species is wrong. Several added phenotype details are unsupported, and the heterodimer partner is incorrectly given as PDGFA instead of PDGFB.
\end{description}

\exlabel{18}
\begin{description}[leftmargin=1.9em,style=nextline,itemsep=1pt,topsep=2pt,font=\normalfont]
\item[Reference:] Dictyostelium discoideum (Social amoeba); FUNCTION: Probable adapter protein and signal transducer that links members of the tumor necrosis factor receptor family to different signaling pathways by association with the receptor cytoplasmic domain and kinases. \{ECO:\allowbreak{}0000250\}.
\item[Prediction:] Dictyostelium discoideum (Social amoeba); FUNCTION: Probable adapter protein and signal transducer that links members of the tumor necrosis factor receptor family to different signaling pathways by association with the receptor cytoplasmic domain and kinases. \{ECO:\allowbreak{}0000250\}.
\item[Expert comment:] Correctly identifies the protein as a heavy-metal-binding zinc/copper protein, but the stated metal composition is wrong and the neuronal inhibitory activity is omitted. The organism is human rather than mouse.
\end{description}

\exlabel{19}
\begin{description}[leftmargin=1.9em,style=nextline,itemsep=1pt,topsep=2pt,font=\normalfont]
\item[Reference:] Xenopus laevis (African clawed frog); FUNCTION: Ligand for members of the frizzled family of seven transmembrane receptors. Plays an important role in embryonic kidney development. Acts downstream of Notch signaling during pronephric kidney development. During early pronephros development, patterns the proximal pronephric anlagen to promote glomus and nephrostome formation. Also required later in pronephros development for tubulogenesis. \{ECO:\allowbreak{}0000269|\allowbreak{}PubMed:\allowbreak{}12142017, ECO:\allowbreak{}0000269|\allowbreak{}PubMed:\allowbreak{}19793883\}.
\item[Prediction:] Mus musculus (Mouse); FUNCTION: Ligand for members of the frizzled family of seven transmembrane receptors. Probable developmental protein. May be a signaling molecule which affects the development of discrete regions of tissues. Is likely to signal over only few cell diameters.
\item[Expert comment:] The molecular identity as a Frizzled ligand is correct, but the organism is wrong and the developmental annotation is overly generic. Key kidney/pronephros-specific roles and Notch-pathway context are missing.
\end{description}

\exlabel{20}
\begin{description}[leftmargin=1.9em,style=nextline,itemsep=1pt,topsep=2pt,font=\normalfont]
\item[Reference:] Arabidopsis thaliana (Mouse-ear cress); FUNCTION: Metallothioneins have a high content of cysteine residues that bind various heavy metals (Probable). Functions as a metal chelator of copper (Cu) and zinc (Zn) (PubMed:18287486). Plays a role in Cu homeostasis, specifically in the remobilization of Cu from senescing leaves. The mobilization of Cu from internal sources is important for seed development (PubMed:24635746). \{ECO:\allowbreak{}0000269|\allowbreak{}PubMed:\allowbreak{}18287486, ECO:\allowbreak{}0000269|\allowbreak{}PubMed:\allowbreak{}24635746, ECO:\allowbreak{}0000305\}.
\item[Prediction:] Arabidopsis thaliana (Mouse-ear cress); FUNCTION: Metallothioneins have a high content of cysteine residues that bind various heavy metals. \{ECO:\allowbreak{}0000305\}.
\item[Expert comment:] Correct species and metallothionein metal-binding function, but the annotation is incomplete. It lacks the specific Cu/Zn chelation and the physiologically important role in Cu remobilization during senescence and seed development.
\end{description}